\documentclass[10pt,twocolumn,letterpaper]{article}

\usepackage{cvpr}            
\usepackage{graphicx}
\usepackage{amsmath}
\usepackage{amssymb}
\usepackage{booktabs}
\usepackage{bm}
\usepackage{multirow}
\usepackage{makecell}

\usepackage[pagebackref,breaklinks,colorlinks]{hyperref}
\usepackage[capitalize]{cleveref}
\crefname{section}{Sec.}{Secs.}
\Crefname{section}{Section}{Sections}
\Crefname{table}{Table}{Tables}
\crefname{table}{Tab.}{Tabs.}

\begin{document}

\title{Deep Image Harmonization with Learnable Augmentation}

\author{Li Niu\thanks{Corresponding author.}~,  Junyan Cao, Wenyan Cong, Liqing Zhang \\
Department of Computer Science and Engineering, MoE Key Lab of Artificial Intelligence, \\
Shanghai Jiao Tong University\\
{\tt \small \{ustcnewly,Joy\_C1\}@sjtu.edu.cn,  wycong@utexas.edu,  zhang-lq@cs.sjtu.edu.cn}
}

\maketitle

\begin{abstract}
The goal of image harmonization is adjusting the foreground appearance in a composite image to make the whole image harmonious. To construct paired training images, existing datasets adopt different ways to adjust the illumination statistics of foregrounds of real images to produce synthetic composite images. However, different datasets have considerable domain gap and the performances on small-scale datasets are limited by insufficient training data. In this work, we explore learnable augmentation to enrich the illumination diversity of small-scale datasets for better harmonization performance. In particular, our designed SYthetic COmposite Network (SycoNet) takes in a real image with foreground mask and a random vector to learn suitable color transformation, which is applied to the foreground of this real image to produce a synthetic composite image. Comprehensive experiments demonstrate the effectiveness of our proposed learnable augmentation for image harmonization. The code of SycoNet is released at \href{https://github.com/bcmi/SycoNet-Adaptive-Image-Harmonization}{https://github.com/bcmi/SycoNet-Adaptive-Image-Harmonization}.
\end{abstract}

\section{Introduction} \label{sec:intro}

\begin{figure*}[t]
\centering
\includegraphics[width=0.86\textwidth]{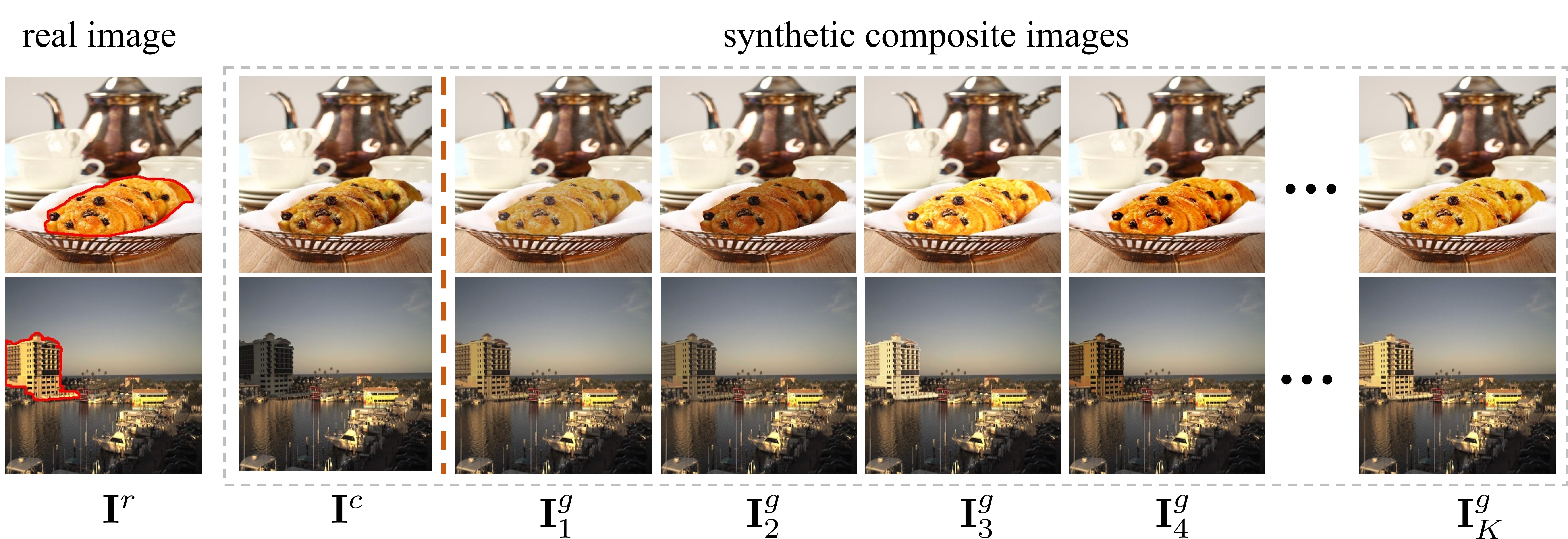} 
\caption{In the left two columns, we show a pair of real image $\mathbf{I}^r$ and original synthetic composite image $\mathbf{I}^c$ from HFlickr (\emph{resp.}, Hday2night) dataset in the top (\emph{resp.}, bottom) row, with the real foreground outlined in red. In the other columns, we show $K$ augmented synthetic composite images $\{\mathbf{I}^g_k|_{k=1}^K\}$ generated by our SycoNet. }
\label{fig:example_composite_image}
\end{figure*}

As a prevalent image editing operation, image composition \cite{niu2021making} aims to cut the foreground from one image and paste it on another background image, making a realistic-looking composite image. Image composition plays a critical role in artistic creation, augmented reality, automatic advertising, and so on \cite{MISCWeng2020,WhatWhereZhang2020}. However, the illumination statistics of foreground and background could be inconsistent due to distinct capture conditions or capture devices (\emph{e.g.}, season, weather, time of the day, camera setting), which makes the resultant composite image unrealistic. To address this issue, image harmonization \cite{CongDoveNet2020} targets at adjusting the illumination statistics of foreground to make it compatible with the background, leading to a harmonious composite image. 

Training data-hungry deep learning models for image harmonization relies on abundant pairs of composite images and harmonious images. Nevertheless, it is extremely difficult and expensive to manually adjust the foreground of composite image to produce harmonious image. Therefore, recent works turn to an inverse manner, that is, adjusting the foreground of real image to produce a synthetic composite image, resulting in pairs of synthetic composite images and ground-truth harmonious real images. In this inverse manner, the work in \cite{CongDoveNet2020} constructed four datasets (HCOCO, HFlickr, HAdobe5k, and Hday2night), which are collectively called iHarmony4. Despite similar construction pipeline, four datasets adopt different foreground adjustment approaches. In particular, HCOCO and HFlickr adopt traditional color transfer methods \cite{reinhard2001color,xiao2006color,fecker2008histogram,pitie2007automated} to adjust the foreground, while HAdobe5k (\emph{resp.}, Hday2night) replaces the foreground with the counterpart retouched by different experts (\emph{resp.}, captured at different times). 
Previous works usually train a deep image harmonization model based on the union of training sets from four datasets. However, the data distributions of different datasets are considerably different, which is caused by many factors (\emph{e.g.}, image source, capture device, scene type, foreground adjustment approach). Following the terminology of domain adaptation~\cite{torralba2011unbiased,patel2015visual}, \textbf{we treat each dataset as one domain and four datasets have large domain gap}. We observe that finetuning on different datasets can bring notable performance gain (see Section \ref{sec:effect_aug}), but the performance is still limited by insufficient training data on small-scale datasets (\emph{e.g.}, HFlickr, Hday2night). 

In this work, we attempt to augment small-scale datasets with more synthetic composite images to enrich the illumination diversity, which may not be easily achieved by using the original foreground adjustment approach. Specifically, for HFlickr, we need to manually filter unqualified synthetic composite images \cite{CongDoveNet2020}, otherwise the performance would be significantly compromised (see Section \ref{sec:effect_aug}). For Hday2night, it is almost impossible to recapture the same scene at different times. Therefore, we design an augmentation network named SYthetic COmposite Network (SycoNet) to produce synthetic composite images automatically, which simulates the original foreground adjustment approach.

\begin{figure*}[t]
\centering
\includegraphics[width=0.9\textwidth]{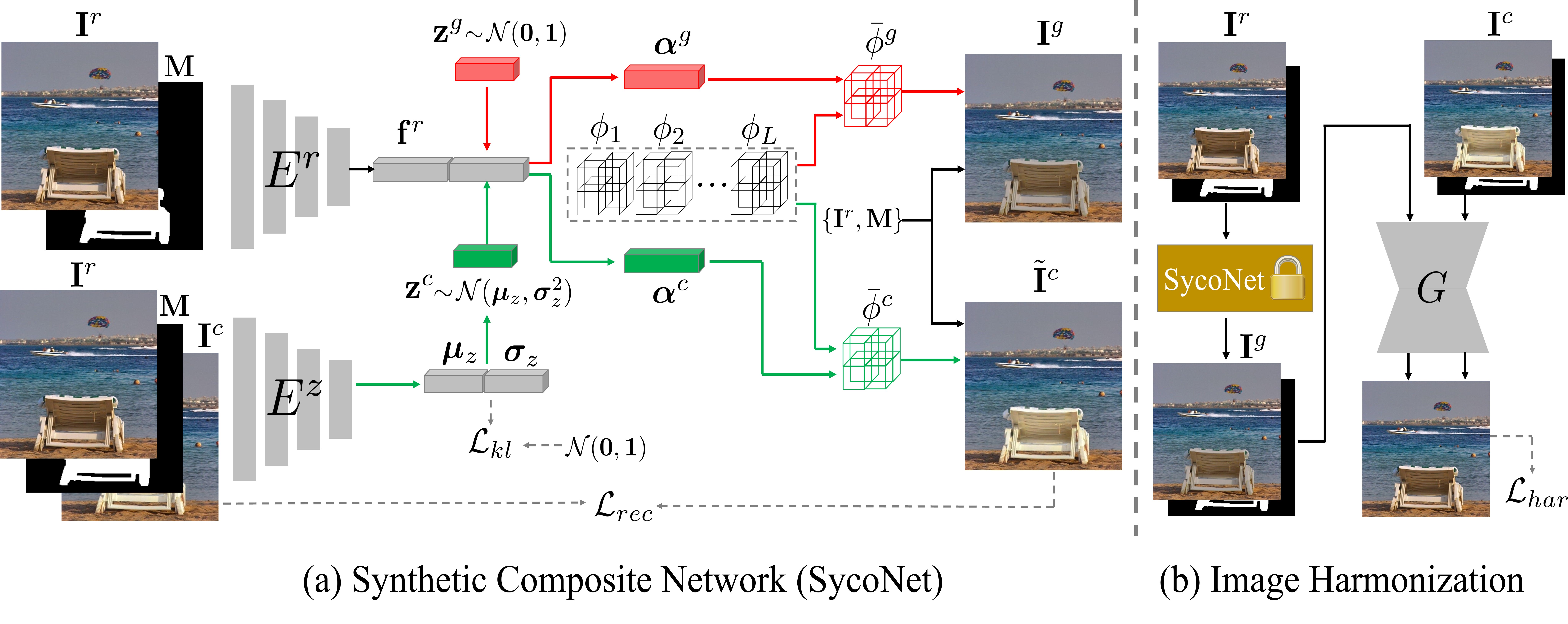} 
\caption{(a) Illustration of our augmentation network SycoNet. In the generation (\emph{resp.}, reconstruction) branch indicated by red (\emph{resp.}, green) arrows, we predict the color transformation LUT $\bar{\phi}^g$ (\emph{resp.}, $\bar{\phi}^c$) to adjust the real foreground in $\mathbf{I}^r$, resulting in the synthetic composite image $\mathbf{I}^g$ (\emph{resp.}, $\tilde{\mathbf{I}}^c$). (b) Image harmonization with augmented synthetic composite images $\mathbf{I}^g$ generated by fixed SycoNet (only generation branch). Both original $\mathbf{I}^c$ and augmented $\mathbf{I}^g$ are harmonized by the harmonization network $G$.}
\label{fig:flowchart}
\end{figure*}

Our SycoNet takes in a real image with foreground mask and a random vector, generating a synthetic composite image with adjusted foreground. By sampling multiple random vectors, we can generate multiple synthetic composite images. As shown in Figure \ref{fig:flowchart}(a), we adopt a CNN encoder to extract feature from real image and foreground mask, which is concatenated with a random vector to produce suitable color transformation for the foreground. The color transformation is realized by a linear combination of basis LUTs \cite{cong2021high}. The combined LUT is applied to the foreground to produce a synthetic composite image. Moreover, we establish a bijection between random vectors and original synthetic composite images in the dataset to ensure the quality and diversity of generated synthetic composite images. Concretely, we employ another CNN encoder to produce a latent code, which is expected to encode the necessary information required to transfer from real foreground to original composite foreground. Thus, when using this latent code to predict color transformation, we hope that the generated synthetic composite image can reconstruct the original synthetic composite image in the dataset.

After training SycoNet, we freeze the model parameters and integrate it with an existing image harmonization network, as shown in Figure \ref{fig:flowchart}(b). When training the image harmonization network, besides the original training pairs of synthetic composite images and real images, SycoNet produces extra synthetic composite images as augmented data. Both original and augmented synthetic composite images (see Figure \ref{fig:example_composite_image}) should be harmonized to approach ground-truth real images. 

\textbf{Our proposed learnable augmentation is helpful for adapting a pretrained image harmonization model to a new domain with limited data.  Given a test image which we do not know which domain it belongs to, we can apply our learnable augmentation in the following two ways.} 1) We use learnable augmentation to enhance the harmonization model of each domain. Given a test image, we can first predict its domain label using a domain classifier and apply the model of the corresponding domain (see Section~\ref{sec:real_composite}). 2) We use learnable augmentation to enhance one unified harmonization model for all domains (see Section~\ref{sec:unified_model}). The second option is more compact, at the cost of performance degradation.  

We conduct experiments on iHarmony4, which demonstrates that our proposed learnable augmentation can significantly boost the harmonization performance.
Our major contributions are summarized as follows: 1) We propose learnable augmentation to enrich the illumination diversity for image harmonization; 2) We design a novel augmentation network named SycoNet, which can automatically generate synthetic composite images by simulating the foreground adjustment in the original dataset; 3) Extensive experiments prove the effectiveness of our proposed learnable augmentation.

\section{Related Work}

\subsection{Image Harmonization}

Traditional image harmonization methods~\cite{colorharmonization,lalonde2007using,xue2012understanding,multi-scale} mainly leveraged traditional color transfer methods (\emph{e.g.}, shifting and scaling,  histogram matching) to align the color information between foreground and background.
In \cite{zhu2015learning}, they designed a model to learn the color shifting matrix by using a discriminator to push the transformed image to be realistic.
Recently, lots of deep image harmonization methods \cite{CongDoveNet2020,sofiiuk2021foreground,Hao2020bmcv,ling2021region,Jiang_2021_ICCV,guo2021image,PHDNet} have emerged, which usually learn a mapping network from composite image to harmonized image. To name a few, \cite{tsai2017deep} supplemented basic image harmonization network with extra semantic information. By treating different capture conditions as different domains, \cite{CongDoveNet2020} proposed to pull close foreground domain and background domain via domain verification, while \cite{bargain} used background domain code to guide the translation of foreground.  \cite{xiaodong2019improving} developed various types of attention blocks embedded in the image harmonization network. \cite{ling2021region} borrowed the idea of adaptive instance normalization from style transfer to image harmonization, which is further extended in \cite{hang2022scs} by considering local style transfer. \cite{guo2021intrinsic} decomposed a composite image to reflectance map and illumination map, followed by modulating the foreground illumination map. \cite{Jiang_2021_ICCV} constructed training pairs based on two crops from the same image and designed an image harmonization model requiring complete background image.  \cite{cong2021high,ke2022harmonizer,liang2021spatial,xue2022dccf} merged color transformation into deep learning networks for efficient image harmonization. 

Different from the above works, we explore learnable augmentation for image harmonization, which has never been studied before. Our designed augmentation network can be integrated with any image harmonization network.

\subsection{Data Augmentation}

Data augmentation targets at augmenting training set with new samples. Traditional data augmentation techniques (\emph{e.g.}, crop, flip, affine transformation, cutout \cite{devries2017improved}) have been widely used in myriads of computer vision tasks. There are also some more advanced augmentation techniques \cite{zhang2018mixup,yun2019cutmix} which mix up training images and their labels. However, all the above augmentation techniques are non-learnable augmentation without learnable parameters. Our proposed learnable augmentation shares some similar thoughts with a research line called auto-augmentation. 
Specifically, AutoAugment~\cite{cubuk2019autoaugment} proposed to search the best data augmentation policy for a target dataset via reinforcement learning, which is accelerated by subsequent works \cite{lim2019fast,tian2020improving,mounsaveng2021learning} using different techniques (\emph{e.g.}, efficient density matching, weight sharing strategy, bilevel optimization). PBA~\cite{ho2019population} 
proposed to learn augmentation policy schedule rather than a fixed policy.
Instead of searching the optimal augmentation policy, we predict proper color transformation to generate augmented data. 

This work makes the first attempt at learnable augmentation for the image harmonization task, in which we generate more synthetic composite images for a real image to construct more training pairs. \textbf{Our augmentation strategy focuses on enriching the illumination diversity of training set, which is complementary with other data augmentation techniques  (\emph{e.g.}, crop, flip, more foregrounds).} We show that enriching the illumination diversity of training set can greatly enhance the image harmonization performance.

\section{Our Method}

We suppose that the training set of an image harmonization dataset contains $N$ pairs of synthetic composite images and real images, \emph{i.e.}, $\mathcal{S}=\{(\mathbf{I}^c_n, \mathbf{I}^r_n)|_{n=1}^N\}$, in which $\mathbf{I}^c_n$ (\emph{resp.}, $\mathbf{I}^r_n$) is the $n$-th synthetic composite image (\emph{resp.}, real image). In the first stage, we train an augmentation network on the training set, which can generate multiple synthetic composite images $\{\mathbf{I}^g_{n,k}|_k\}$ for a real image $\mathbf{I}^r_n$. In the second stage, we integrate the augmentation network into an existing image harmonization network $G$ and fix the model parameters of augmentation network. When training the image harmonization network, the augmentation network dynamically generates augmented synthetic composite images $\{\mathbf{I}^g_{n,k}|_{n=1,k}^N\}$  to supplement the original training set  $\mathcal{S}$, aiming to train a better image harmonization model. Next, we will introduce the first stage in Section~\ref{sec:syco} and the second stage in Section~\ref{sec:aug}.

\subsection{Synthetic Composite Network}\label{sec:syco}
We refer to our augmentation network as SYnthetic COmposite Network (SycoNet), which can generate multiple synthetic composite images given a real image. Our SycoNet consists of a generation branch and a reconstruction branch, which will be detailed separately. In this section, we omit the subscript $n$ for brevity. Besides, the foreground of $\mathbf{I}^r$ (\emph{resp.}, $\mathbf{I}^c$, $\mathbf{I}^g$) is denoted as $\mathbf{F}^r$ (\emph{resp.}, $\mathbf{F}^c$, $\mathbf{F}^g$).

\subsubsection{Generation Branch} \label{sec:generation_branch}
In the generation branch, we produce suitable color transformation function $h(\cdot)$ for the foreground $\mathbf{F}^r$ of real image $\mathbf{I}^r$, in which $h(\cdot)$ could convert the source color value $\mathbf{v}_{src}$ in $\mathbf{F}^r$ to a target color value $\mathbf{v}_{tgt}=h(\mathbf{v}_{src})$. The color transformation function $h(\cdot)$ can be realized in various forms. In this work, we opt for look-up table (LUT), which has been widely used in a variety of computer vision tasks \cite{2006Image,2015Randomized,2011Medical,Zeng2020,cong2021high}. 
Briefly speaking, a look-up table (LUT) is a 3D lattice in the RGB color space with each dimension corresponding to one color channel (\emph{e.g.}, red). An LUT has $(B + 1)^3$ entries by uniformly slicing the color space into $B$ bins in each dimension, where $B$ is set as 16 following \cite{Zeng2020,cong2021high}. Each entry in the LUT has an indexing color and its output color. Given a source color value $\mathbf{v}_{src}$, its target color value $\mathbf{v}_{tgt}$ could be obtained by looking up its eight nearest indexing colors in the LUT and performing trilinear interpolation based on their output colors. More technical details of LUT can be found in \cite{Zeng2020,cong2021high}.

Following \cite{Zeng2020,cong2021high}, we learn a group of $L$ basis LUTs $\{\phi_l|_{l=1}^L\}$ shared among all images and predict image-specific combination coefficients of basis LUTs for each image. In this way, shared basis LUTs are combined adaptively to form image-specific LUT. In \cite{Zeng2020,cong2021high}, $L$ basis LUTs are initialized as one identity map and $L\!-\!1$ zero maps. Besides, there is no constraint for the combination coefficients, that is, the  coefficient can be either positive or negative.
The $L\!-\!1$ LUTs initialized with zero maps actually function as residual LUTs  \cite{cong2021high}. However, during our experiments, we observe that the learnt residual LUTs are prone to be similar with each other, which severely limits the representation ability of combined LUT. Therefore, we modify the LUT initialization and combination strategy. Precisely, we initialize $L$ basis LUTs with one identity LUT and $L\!-\!1$ representative LUTs (we collect 100 LUTs from Internet and perform clustering to get $L\!-\!1$ cluster centers).  All $L$ basis LUTs are updated during training. Moreover, the combination coefficients are softmax normalized, so that all coefficients are positive and sum up to one. The comparison results demonstrate the advantage of our LUT initialization and combination strategy (see Table \ref{tab:ablation_study}). 

After introducing the color transformation function $h(\cdot)$ realized in the form of LUT, we describe the network architecture of generation branch.  
Given a real image $\mathbf{I}^r$ and its foreground mask $\mathbf{M}$, we concatenate them and feed into an encoder $E^r$ (\emph{e.g.}, ResNet18 \cite{he2015resnet}) to produce a feature vector $\mathbf{f}^r$. For each real image, we hope to generate multiple synthetic composite images instead of a single deterministic one. As a common approach to support stochastic sampling, we sample a random vector $\mathbf{z}^g$ from unit Gaussian distribution $\mathcal{N}(\mathbf{0},\mathbf{1})$ and concatenate it with $\mathbf{f}^r$. Then, the concatenation $[\mathbf{f}^r,\mathbf{z}^g]$ passes through one fully-connected (FC) layer followed by softmax normalization to produce the combination coefficients $\bm{\alpha}^g$. Given the learnt basis LUTs $\{\phi_l|_{l=1}^L\}$, the combined LUT is $\bar{\phi}^g=\sum_{l=1}^L \alpha^g_l \phi_l$, in which $\alpha^g_l$ is the $l$-th element in $\bm{\alpha}^g$. Finally, we apply $\bar{\phi}^g$ to $\mathbf{F}^r$ and get the transformed foreground $\mathbf{F}^g$, which is combined with original background to compose a synthetic composite image $\mathbf{I}^g$.

\subsubsection{Reconstruction Branch} \label{sec:reconstruction}
Note that the generation branch ignores the original synthetic composite images $\mathbf{I}^c$ in the training set $\mathcal{S}$ and lacks supervision for the generated synthetic composite images  $\mathbf{I}^g$. Hence, there is no guarantee for the plausibility of generated synthetic composite images.
In the reconstruction branch, we aim to ensure the quality and diversity of generated synthetic composite images, by
reconstructing original synthetic composite image $\mathbf{I}^c$ from real image $\mathbf{I}^r$. 
 
Given the original synthetic composite image $\mathbf{I}^c$ for real image $\mathbf{I}^r$, we deliver the concatenation of $\mathbf{I}^c$, $\mathbf{I}^r$, and $\mathbf{M}$ to an encoder $E^z$ (\emph{e.g.}, ResNet18 \cite{he2015resnet}). $E^z$ produces $\bm{\mu}_z$ and $\bm{\sigma}_z$, based on which the latent code $\mathbf{z}^c$ could be sampled from Gaussian distribution $\mathcal{N}(\bm{\mu}_z, \bm{\sigma}_z^2)$. By using reparameterization trick \cite{kingma2013auto}, we obtain $\mathbf{z}^c= \bm{\mu}_z+\bm{\epsilon}\odot \bm{\sigma}_z$, where $\bm{\epsilon}$ is a random vector sampled from $\mathcal{N}(\mathbf{0},\mathbf{1})$ and $\odot$ means element-wise product.
$\mathbf{z}^c$ is expected to encode the requisite information to generate $\mathbf{I}^c$ conditioned on $\mathbf{I}^r$.
Analogous to the generation branch, we concatenate $\mathbf{z}^c$ with $\mathbf{f}^r$, and use the same FC layer to produce the combination coefficients $\bm{\alpha}^c$ of basis LUTs. Then, we apply the combined LUT $\bar{\phi}^c=\sum_{l=1}^L \alpha^c_l \phi_l$ to the real foreground $\mathbf{F}^r$ and get a synthetic composite image $\tilde{\mathbf{I}}^c$, which is pushed towards $\mathbf{I}^c$ using the reconstruction loss $\mathcal{L}_{rec} = \|\tilde{\mathbf{I}}^c-\mathbf{I}^c\|_1$. In the meanwhile, we use KL divergence loss \cite{kullback1951information} $\mathcal{L}_{kl}=KL[\mathcal{N}(\bm{\mu}_z, \bm{\sigma}_z^2)||\mathcal{N}(\mathbf{0},\mathbf{1})]$ to enforce $\mathcal{N}(\bm{\mu}_z, \bm{\sigma}_z^2)$ to approach $\mathcal{N}(\mathbf{0},\mathbf{1})$. 

The above reconstruction process establishes a bijection between latent code and synthetic composite image conditioned on real image, which enables generating qualified and diverse synthetic composite images by sampling $\mathbf{z}^g\sim \mathcal{N}(\mathbf{0}, \mathbf{1})$. The reasons are explained as follows. In terms of quality, $\mathbf{I}^r$ and $\mathbf{z}^c$ can produce one plausible color transformation (\emph{i.e.}, from $\mathbf{I}^r$ to $\mathbf{I}^c$) due to the bijection. We assume that such knowledge can be transferred across the joint space of real image and latent code, so that  $\mathbf{I}^r$ and other sampled $\mathbf{z}^g$ can produce other plausible color transformations. 
More specifically, we denote that $\mathbf{I}^r_i$ and $\mathbf{z}_i^c$ can reconstruct plausible composite image $\mathbf{I}_i^c$ for $i=1,\ldots,N$. Assuming that $\{\mathbf{z}_i^c|_{i=1}^N\}$ are transferrable across real images $\{\mathbf{I}^r_i|_{i=1}^N\}$, $\mathbf{I}^r_i$ and $\mathbf{z}_j^c$ for $j\neq i$ could also produce plausible composite images. Besides, $\{\mathbf{z}_i^c|_{i=1}^N\}$ are sampled from Gaussian distributions close to $\mathcal{N}(\mathbf{0},\mathbf{1})$, so  $\mathbf{I}^r_i$ and $\mathbf{z}^g\sim \mathcal{N}(\mathbf{0},\mathbf{1})$ could also produce plausible composite images.
In terms of diversity, the bijection can prevent two latent codes from producing the same synthetic composite image. If two latent codes produce the same synthetic composite image, one synthetic composite image cannot be mapped back to two different latent codes, which would violate the bijection. The experiments in Table \ref{tab:ablation_study} verify that the reconstruction branch can effectively promote the quality and diversity of generated synthetic composite images.

So far, the total loss function can be summarized as
\begin{eqnarray}\label{eqn:syco_loss}
\mathcal{L}_{syco} = \mathcal{L}_{kl}+\mathcal{L}_{rec}.
\end{eqnarray}
After training the augmentation network on the training set $\mathcal{S}$ using Eqn. (\ref{eqn:syco_loss}), the obtained augmentation network could generate more synthetic composite images to augment the original training set. 
Formally, given the $n$-th real training image $\mathbf{I}^r_n$, we can produce $K$ synthetic composite images $\{\mathbf{I}^g_{n,k}|_{k=1}^K\}$ by sampling $K$ random vectors $\{\mathbf{z}^g_k|_{k=1}^K\}$. 
Recall that different datasets can be treated as  different domains with different data distributions, it would be beneficial to train the augmentation network on each dataset separately (see Section~\ref{sec:effect_aug}). When training on a specific dataset, the reconstruction branch could simulate the foreground adjustment process of this dataset, and the generation branch could produce synthetic composite images with close data distribution to the original ones in this dataset.

\subsection{Dynamic Augmentation}\label{sec:aug}

With the trained SycoNet in Section~\ref{sec:syco}, we can integrate its generation branch with any existing image harmonization network $G$ for dynamic augmentation, as illustrated in Figure \ref{fig:flowchart}. Normally, $G$ is trained on the training set $\mathcal{S}=\{(\mathbf{I}^c_n, \mathbf{I}^r_n)|_{n=1}^N\}$, by harmonizing $\mathbf{I}^c_n$ to be close to $\mathbf{I}^r_n$. In our training process, for the $n$-th real training image $\mathbf{I}^r_n$ in the $t$-th training iteration, we use fixed SycoNet to produce a synthetic composite image $\mathbf{I}^g_{n,t}$ with sampled  $\mathbf{z}^g_t\sim \mathcal{N}(\mathbf{0}, \mathbf{1})$. We denote the harmonization results of $\mathbf{I}^c_n$ and $\mathbf{I}^g_{n,t}$ as $\tilde{\mathbf{I}}^c_n$ and $\tilde{\mathbf{I}}^g_{n,t}$ respectively, both of which should be close to $\mathbf{I}^r_n$. The loss function used in original image harmonization network $G$ is represented by $\mathcal{L}_{har}$ and may be different in various image harmonization methods. So we omit the details of $\mathcal{L}_{har}$ here. Then, the training loss can be written as
\begin{eqnarray}\label{eqn:loss_har}
\mathcal{L}_{har} = \sum_{n=1}^N \sum_{t=1}^T \mathcal{L}_{har}(\tilde{\mathbf{I}}^c_n,\mathbf{I}^r_n) + \mathcal{L}_{har}(\tilde{\mathbf{I}}^g_{n,t},\mathbf{I}^r_n),
\end{eqnarray} 
in which $T$ is the number of training iterations. Since the augmented composite images dynamically vary during the training procedure, we refer to this augmentation strategy as dynamic augmentation. We also compare this augmentation strategy with static augmentation, that is, generating adequate augmented composite images beforehand and merging them into the original training set. We find that dynamic augmentation is more elegant and effective than static augmentation (see Section \ref{sec:cmp_dynamic_static}). 

Note that SycoNet is only used in the training stage. During inference, we only use the image harmonization network $G$, without introducing extra computational cost. 

\begin{table*}[t]
\begin{center}

    \begin{subtable}{1\linewidth}
        \centering
        \begin{tabular}{|c|c|c|c|c|c|c|c|c|}
        \hline
        \multirow{2}{*}{} & \multirow{2}{*}{Train} & \multirow{2}{*}{Aug} &  \multicolumn{3}{c|}{Hday2night} & \multicolumn{3}{c|}{HFlickr} \\  
        \cline{4-9}
        ~ & ~ & ~ & MSE$\downarrow$ & fMSE$\downarrow$ & fSSIM $\uparrow$ &  MSE$\downarrow$ & fMSE$\downarrow$ & fSSIM $\uparrow$ \\ \hline
        1 & - & - & 40.59 & 591.07 & 0.7726 & 69.68 & 443.63 & 0.9108 \\ 
\hline
        2 & ft(o) & - & 38.08 & 567.21 & 0.7763 & 64.07 & 411.12 & 0.9151 \\ 
\hline
   		3 & ft(oa) & CT & 39.62 & 575.35 & 0.7672 & 73.00 & 461.90 & 0.9072 \\ 
\hline
		4 & ft(oa) & LUT & 42.97 & 630.55 & 0.7661 & 86.06 & 559.67 & 0.8948 \\ 
\hline
		5 & ft(oa) & Ours & 36.49 & 547.63 & 0.7766 & 63.53 & 407.53 & 0.9164 \\ 
\hline
		6 & ft(a) & Ours(ft) & 42.79 & 572.53 & 0.7757 & 68.05 & 431.20 & 0.9149 \\ 
\hline
		7 & ft(oa) & Ours(ft) & \textbf{34.44} & \textbf{517.00} & \textbf{0.7844} & \textbf{58.46} & \textbf{385.43} & \textbf{0.9176} \\ 
\hline
       \end{tabular}
        \caption{The results using iS$^2$AM \cite{sofiiuk2021foreground} as the harmonization network.}
        \label{tab:data_augmentation_a}
    \end{subtable}
    
    \vfill

    \begin{subtable}{1\linewidth}
        \centering
        \begin{tabular}{|c|c|c|c|c|c|c|c|c|}
		\hline
		\multirow{2}{*}{} & \multirow{2}{*}{Train} & \multirow{2}{*}{Aug} & \multicolumn{3}{c|}{Hday2night} & \multicolumn{3}{c|}{HFlickr} \\  
\cline{4-9}
		~ & ~ & ~ & MSE$\downarrow$ & fMSE$\downarrow$ & fSSIM $\uparrow$ &  MSE$\downarrow$ & fMSE$\downarrow$ & fSSIM $\uparrow$ \\ \hline
		1 & - & - & 47.24 & 852.12 & 0.7444 & 117.59 & 751.13 & 0.8573 \\ 
\hline
		2 & ft(o) & - & 44.10 & 785.79 & 0.7471 & 107.83 & 722.97 & 0.8629 \\ 
\hline
		3 & ft(oa) & CT & 51.10 & 950.66 & 0.7177 & 105.69 & 708.03 & 0.8685 \\ 
\hline
		4 & ft(oa) & LUT & 50.44 & 937.03 & 0.7349 & 117.72 & 788.21 & 0.8533 \\ 
\hline
		5 & ft(oa) & Ours & 39.86 & 717.05 & 0.7531 & 102.57 & 678.18 & 0.8731 \\ 
\hline
		6 & ft(a) & Ours(ft) & 44.47 & 792.99 & 0.6758 & 109.94 & 733.49 & 0.8614 \\ 
\hline
		7 & ft(oa) & Ours(ft) & \textbf{37.28} & \textbf{643.78} & \textbf{0.7662} & \textbf{96.01} & \textbf{634.55} & \textbf{0.8780} \\ 
\hline
		\end{tabular}
        \caption{The results using RainNet~\cite{ling2021region} as the harmonization network.}
        \label{tab:data_augmentation_c}
    \end{subtable}

    \vfill
    
    \caption{In ``Train" column, ``ft" means finetuning the harmonization model on Hday2night/HFlickr, and ``o" (\emph{resp.}, ``a") represents original (\emph{resp.}, augmented) training images. In ``Aug" column, ``LUT" (\emph{resp.}, ``CT") means using random LUT (\emph{resp.}, traditional color transfer methods) for data augmentation. ``Ours" means using our SycoNet for data augmentation, and ``Ours(ft)" means finetuning SycoNet on Hday2night/HFlickr. The best results are highlighted in boldface.}
    \label{tab:data_augmentation}
\end{center}
\end{table*}

\section{Experiments}

\subsection{Datasets and Implementation Details}

We conduct experiments on iHarmony4~\cite{CongDoveNet2020} which consists of four datasets: HCOCO, HFlickr, HAdobe5k, and Hday2night. 
HCOCO (\emph{resp.}, HAdobe5k, HFlickr, and Hday2night) has  38545 (\emph{resp.}, 19437, 7449, and 311) training images and 4283 (\emph{resp.}, 2160, 828, and 133) test images. We mainly use HFlickr and Hday2night, which are two relatively small datasets, because the advantage of data augmentation could be exhibited more clearly on small-scale datasets. We also conduct similar experiments on the other two datasets (HCOCO and HAdobe5k) and observe that the performance gain decreases as the scale of training data increases, which is left to Supplementary.

In the augmentation network, we adopt ResNet18 \cite{he2015resnet} as the encoders $E^r$ and $E^z$. We set the dimension of latent code $\mathbf{z}$ as $d_z=32$ and the number of basis LUTs as $L=20$. For the harmonization network $G$, since our augmentation network can cooperate with any existing image harmonization network, we take RainNet \cite{ling2021region} and iS$^2$AM \cite{sofiiuk2021foreground} as two examples to show the effectiveness of learnable augmentation in Section \ref{sec:effect_aug}.
In the other sections, we adopt iS$^2$AM as $G$ by default considering its simplicity and effectiveness.  
Our network is implemented using Pytorch 1.7.0 and trained using Adam optimizer with learning rate of $1e{-4}$ on ubuntu 18.04 LTS operation system, with 32GB memory, Intel Core i7-9700K CPU, and two GeForce GTX 2080 Ti GPUs. 
We adopt MSE, foreground MSE (fMSE), foreground SSIM (fSSIM) as evaluation metrics for harmonization performance, in which MSE is calculated based on the whole image while fMSE and fSSIM are calculated within the foreground region.   

\subsection{Effectiveness of Our Learnable Augmentation} \label{sec:effect_aug}

\begin{figure*}[t]
\centering
\includegraphics[width=0.99\textwidth]{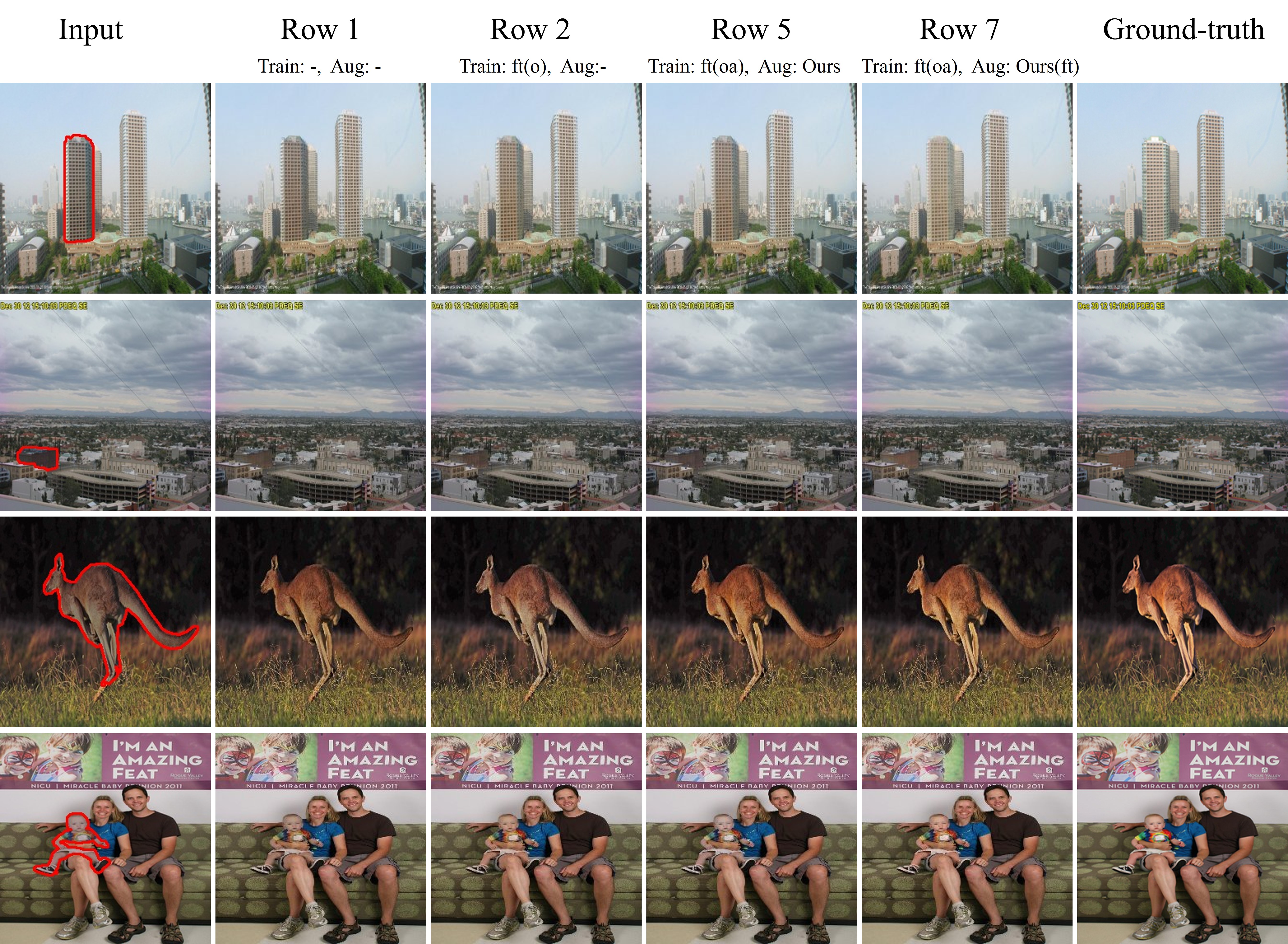} 
\caption{The leftmost (\emph{resp.}, rightmost) column is the input composite image (\emph{resp.}, ground-truth real image), in which the foreground in the input image is outlined in red. The rest columns show the harmonized results of row 1, 2, 5, 7 in Table \ref{tab:data_augmentation_a}. \emph{Train} and \emph{Aug} below the row index means the corresponding training setting and augmentation setting as in Table \ref{tab:data_augmentation}. The first (\emph{resp.}, last) two rows are from Hday2night (\emph{resp.}, HFlickr) dataset. }
\label{fig:example_results}
\end{figure*}

In Table \ref{tab:data_augmentation_a}, we first report the results of iS$^2$AM model trained on the whole iHarmony4 training set (row 1). Then, we finetune the harmonization model on the training set of Hday2night or HFlickr (``ft(o)"). The attained results in row 2 are significantly better than those in row 1, which verifies the domain gap between different datasets. Next, the following results are all obtained by finetuning the harmonization model on Hday2night or HFlickr, with both original training data and augmented training data (``ft(oa)") or only augmented training data (``ft(a)").

We first try two simple augmentation methods: traditional color transfer methods \cite{reinhard2001color,xiao2006color,fecker2008histogram,pitie2007automated} (``CT" in row 3) and random LUT (``LUT" in row 4). For ``CT", following  \cite{CongDoveNet2020}, for each real image in each training iteration, we randomly choose one reference image from ADE20k dataset~\cite{zhou2019semantic} and one of the color transfer methods \cite{reinhard2001color,xiao2006color,fecker2008histogram,pitie2007automated} to perform color transfer on its foreground. The attained results in row 3 are worse than those in row 2. As claimed in \cite{CongDoveNet2020}, the obtained synthetic composite images without deliberate filtering may have noticeable artifacts or unreasonable albedo change, which would adversely affect the effectiveness of augmentation. For ``LUT", we gather 100 LUTs from Internet and randomly choose one LUT for each real image in each training iteration. The obtained results in row 4 become even worse. 

Then, we train our augmentation network SycoNet on the whole iHarmony4 training set and perform dynamic augmentation when finetuning the harmonization model. The obtained results in row 5 outperform two simple augmentation methods (row 3 and row 4), and generally outperform those without augmentation (row 2),  which shows the superiority of our augmentation network. Next, we finetune SycoNet on Hday2night and HFlickr respectively, and use finetuned SycoNet for dynamic augmentation. We report the results in the last row (row 7), based on which the finetuned SycoNet performs more favorably (row 7 \emph{v.s.} row 5). Due to the large domain gap between different datasets, the finetuned SycoNet can better approximate the foreground adjustment process of each dataset, and thus generate more suitable synthetic composite images for each dataset. Finally, we report the results (row 6) obtained by only using augmented data and discarding original training data. The comparison between row 6 and row 2 implies that the generated synthetic composite images still have certain gap with the original synthetic composite images. However, jointly using them can achieve significant improvement (row 7 \emph{v.s.} row 2).
Another observation is that the improvement of our learnable augmentation (row 7 \emph{v.s.} row 2) on Hday2night dataset is more significant than that on HFlickr dataset, which confirms our conjecture that data augmentation is more effective on small-scale datasets. 

In Table  \ref{tab:data_augmentation_c}, we report the results based on RainNet and have similar observations on the relation between different rows. For example, simple color augmentation (row 3, row 4) generally brings no performance gain, except when the performance without augmentation is very poor (\emph{e.g.}, ``CT" slightly improves RainNet on HFlickr). A common SycoNet can improve the results (row 5 \emph{v.s.} row 2) and the finetuned SycoNet can achieve further improvement (row 7 \emph{v.s.} row 5). 
The improvement on Hday2night is more notable than that on HFlickr.  


\subsection{Qualitative Results}

For qualitative comparison, we show the visualization results of row 1, 2, 5, 7 in Table \ref{tab:data_augmentation_a} on two datasets in Figure \ref{fig:example_results}. From row 1 to row 7, the results are overall getting better. The results obtained by using our finetuned augmentation network (row 7) are more visually appealing and closer to the ground-truth real images, which proves that it is useful to finetune the harmonization model with the aid of finetuned augmentation network. More visualization results can be found in Supplementary.

\subsection{Ablation Studies on Augmentation Network}\label{sec:ablation_study}

By taking Hday2night as an example, we conduct ablation studies on our augmentation network SycoNet in Table \ref{tab:ablation_study}.
Besides the harmonization metrics used in Table \ref{tab:data_augmentation}, we also evaluate the diversity of generated composite images, which is referred to as ``Div" in Table \ref{tab:ablation_study}. In particular, for each real training image, we randomly sample $\mathbf{z}^g$ for 10 times to produce 10 synthetic composite images. Then, we calculate fMSE between each pair of composite images and compute the average over all pairs, and then compute the average over all real training images. For MSE, fMSE, fSSIM, the experimental setting is the same as row 7 in Table \ref{tab:data_augmentation_a}. In detail, we finetune the harmonization model on each dataset using data augmentation, with the augmented images generated by different variants of our augmentation network. We include the results obtained by using our full-fledged augmentation network (row 7 in Table \ref{tab:data_augmentation_a}) as ``Ours" for comparison.

First, we delete the reconstruction branch by removing $E^z$ and the associated inputs/outputs. After removal, there is no supervision for the generated synthetic composite image $\mathbf{I}^g$. Therefore, we add an adversarial loss to make $\mathbf{I}^g$ indistinguishable from original synthetic composite images $\mathbf{I}^c$ in the training set. We observe that the produced combination coefficients collapse to an one-hot vector and the generated synthetic composite images lack diversity, which is known as mode collapse issue~\cite{zhu2017toward}. As reported in row 1, ``Div" is close to zero and the effect of augmentation is negligible compared with row 2 in Table \ref{tab:data_augmentation_a}.   
We also try removing $\mathbf{I}^r$ from the input of $E^z$, because only using $\{\mathbf{I}^c, \mathbf{M}\}$ could also provide the target illumination information of composite foreground $\mathbf{F}^c$. The results become worse than ``Ours" (row 2 \emph{v.s.} row 6), which shows that it would be better to use both $\mathbf{I}^r$ and  $\mathbf{I}^c$ to encode the information required to transfer from $\mathbf{F}^r$ to $\mathbf{F}^c$.
Considering that the random vector $\mathbf{z}$ could be appended to different locations in $E^z$, we try an alternative location, \emph{i.e.}, spatially replicating $\mathbf{z}$ and appending it to the input of $E^z$. The attained results become slightly worse than ``Ours" (row 3 \emph{v.s.} row 6). 
Recall that we make some modifications about the LUT design compared with \cite{Zeng2020,cong2021high}. Here, we change the LUT design to be the same as in \cite{Zeng2020,cong2021high} and report the results in row 4. As mentioned in Section~\ref{sec:generation_branch}, the learnt residual LUTs are prone to be similar with each other, so the diversity of generated composite images is degraded. 
The harmonization performance is also worse than ``Ours", indicating that our LUT design is more effective. We also try sampling $\bm{z}^c$ to generate composite images and report the results in row 6. Sampling $\bm{z}^c$ can only reconstruct original synthetic composite images (see Section~\ref{sec:reconstruction}) and cannot achieve the goal of augmentation, so the obtained results are merely comparable with those without using augmentation.

\subsection{Comparing Dynamic and Static Augmentation} \label{sec:cmp_dynamic_static}

As described in Section~\ref{sec:aug}, we dynamically generate augmented composite images during the training procedure, which is dubbed as dynamic augmentation. There exists another straightforward augmentation strategy: generating augmented composite images beforehand and merging them into the original training set,  which is dubbed as static augmentation.
We compare dynamic augmentation with static augmentation in Supplementary. 

\begin{table}[t]
\centering
\begin{tabular}{|c|c|c|c|c|c|}
\hline
 & Method & MSE$\downarrow$ & fMSE$\downarrow$ & fSSIM $\uparrow$  & Div$\uparrow$\\ \hline
1 & w/o rec & 40.72 & 569.49 & 0.7754  & 0.01  \\ 
\hline
2 & w/o $\mathbf{I}^r$ & 37.64 & 554.48 & 0.7760  & 1094.57  \\ 
\hline
3 & move $\mathbf{z}$ & 34.83 & 525.54 & 0.7834  & 987.47  \\ 
\hline
4 & LUTv2 & 37.10 & 541.56 & 0.7810  & 847.01  \\ 
\hline
5 & $\mathbf{z}^c$ & 37.59 & 556.44 & 0.7758  & 8.37   \\ 
\hline
6 & Ours & 34.44 & 517.00 & 0.7844  & 1132.71  \\ 
\hline
\end{tabular}
\caption{Ablation studies of our data augmentation network SycoNet on Hday2night. ``Div" measures the diversity of augmented images. ``w/o rec" means removing the reconstruction branch. ``w/o $\mathbf{I}^r$" means removing $\mathbf{I}^r$ from $E^z$ input. ``move $\mathbf{z}$" means moving $\mathbf{z}$ to $E^z$ input. ``LUTv2" means using the LUT design in \cite{Zeng2020,cong2021high}. $\mathbf{z}^c$ means using $\mathbf{z}^c \sim \mathcal{N}(\mathbf{\mu}_z,\mathbf{\sigma}_z^2)$ instead of $\mathbf{z}^g \sim \mathcal{N}(\mathbf{0},\mathbf{1})$ to generate $\mathbf{I}^g$.}
\label{tab:ablation_study}
\end{table}

\subsection{Evaluation on Real Composite Images} \label{sec:real_composite}

Following previous image harmonization works~\cite{tsai2017deep,CongDoveNet2020,ling2021region,sofiiuk2021foreground}, we also evaluate our method on 199 real composite images from \cite{tsai2017deep,cong2021high} (99 images from \cite{tsai2017deep} and 100 images from \cite{cong2021high}).
As there are no ground-truth harmonious images for real composite images, we conduct user study for comparison. The detailed user study results and visualization results are left to Supplementary.

\subsection{One Unified Model to Rule All Domains} \label{sec:unified_model}

We use SycoNet pretrained on iHarmony4 to generate augmented images for the whole iHarmony4 training set. Then, we train one unified model on the augmented training set, which is applied to all four domains. The detailed results and analyses are left to Supplementary. 

\section{Conclusion}
In this work, we have proposed learnable augmentation to enrich the illumination diversity of image harmonization datasets. Specifically, we design a novel augmentation network named SycoNet, which can produce more synthetic composite images for a real image. 
Our SycoNet can be integrated into any existing image harmonization network for dynamic augmentation. 
Comprehensive experiments show that our proposed learnable augmentation can significantly boost the harmonization performance. 

\section*{Acknowledgement} The work was supported by the National Natural Science Foundation of China (Grant No. 62076162), the Shanghai Municipal Science and Technology Major/Key Project, China (Grant No. 2021SHZDZX0102, Grant No. 20511100300). 

{\small
\bibliographystyle{ieee_fullname}
\bibliography{camera_main.bbl}
}

\end{document}


\title{Supplementary for Deep Image Harmonization with Learnable Augmentation}

\author{Li Niu\thanks{Corresponding author.}~,  Junyan Cao, Wenyan Cong, Liqing Zhang \\
Department of Computer Science and Engineering, MoE Key Lab of Artificial Intelligence, \\
Shanghai Jiao Tong University\\
{\tt \small \{ustcnewly,Joy\_C1\}@sjtu.edu.cn,  wycong@utexas.edu,  zhang-lq@cs.sjtu.edu.cn}
}

\maketitle

\begin{figure*}[t]
\centering
\includegraphics[width=0.8\textwidth]{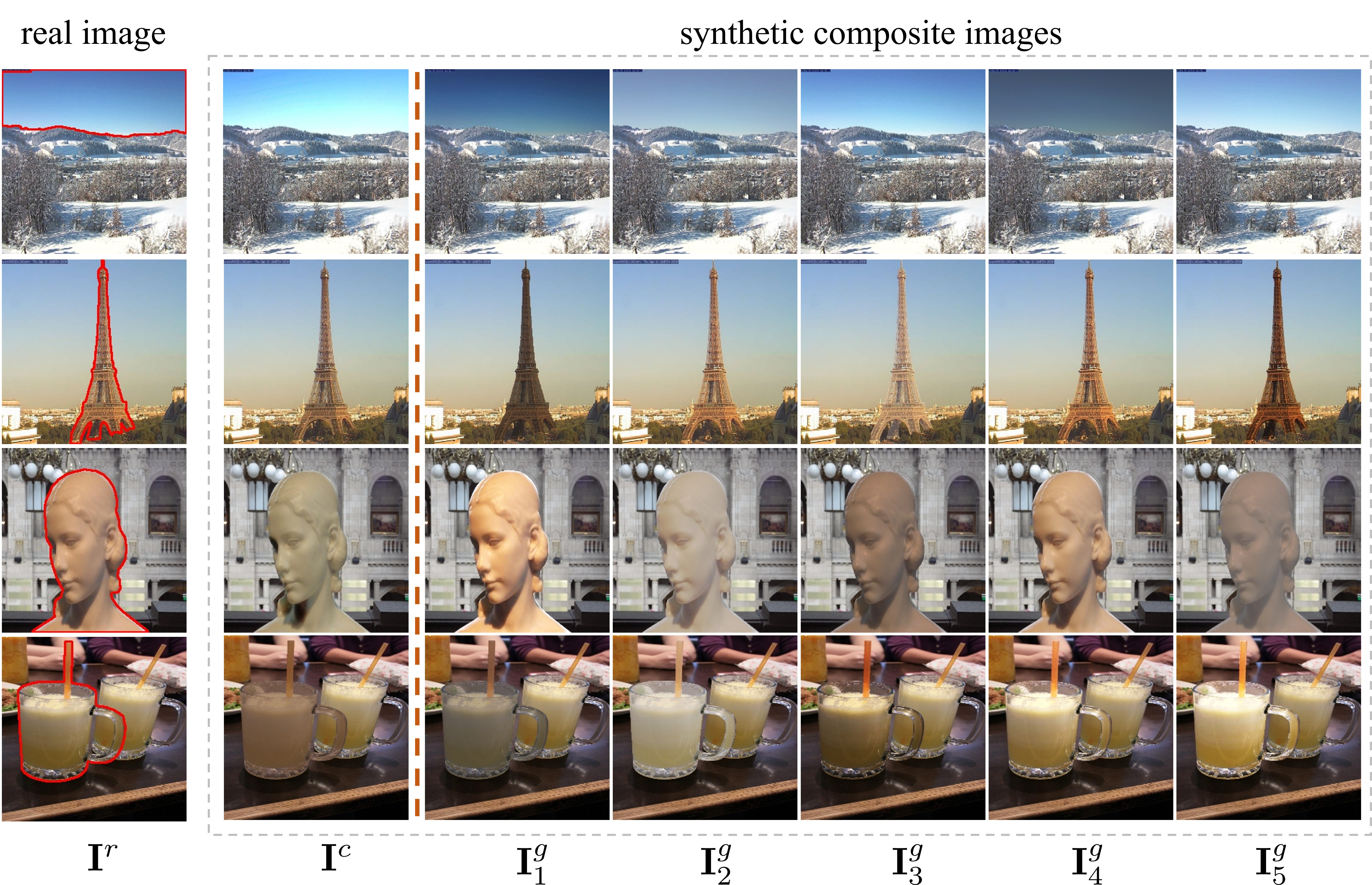} 
\caption{Examples of real images and synthetic composite images. In the left two columns, we show a pair of real image $\mathbf{I}^r$ and synthetic composite image $\mathbf{I}^c$ from Hday2night (\emph{resp.}, HFlickr) dataset in the first (\emph{resp.}, last) two rows, with the real foreground outlined in red. In the rest columns, we show $5$ synthetic composite images $\{\mathbf{I}^g_k|_{k=1}^5\}$ generated by our augmentation network SycoNet. }
\label{fig:supp_example_composite_image}
\end{figure*}

In this document, we provide additional materials to support our main paper. In Section \ref{sec:visual_results}, we will show the synthetic composite images generated by our SycoNet and the harmonized results of different models. 
In Section \ref{sec:large_scale}, we will report the results on HCOCO and HAdobe5k datasets. 
In Section \ref{sec:real_composite}, we will show the harmonized results and user study results on real composite images. 
In Section \ref{sec:cmp_dynamic_static}, we will compare two augmentation strategies: dynamic augmentation and static augmentation. 
In Section \ref{sec:unified_model}, we train one unified model for all four domains with learnable augmentation. 
In Section \ref{sec:hyper_parameter}, we will investigate the impact of two hyper-parameters $d_z$ and $L$.  
In Section \ref{sec:limitation}, we will show some failure cases and provide in-depth analyses. 
In Section \ref{sec:computation_cost}, we will analyze the computational cost of our method.

\section{More Visualization Results}
\label{sec:visual_results}

First, we show the synthetic composite images generated by our SycoNet following the experimental setting in row 7 in Table 1(a) of the main paper. Specifically, we train SycoNet on the whole iHarmony4 training set and finetune SycoNet on Hday2night/HFlickr. Then, we apply the finetuned SycoNet to generate more synthetic composite images based on the real training images in Hday2night/HFlickr. We have shown several examples in Figure 1 in the main paper. Here, we provide more examples in Figure \ref{fig:supp_example_composite_image}. It can be seen that our SycoNet is capable of generating diverse synthetic composite images of high quality for different foreground objects and background scenes. 

\begin{figure*}
\centering
\includegraphics[width=0.85\textwidth]{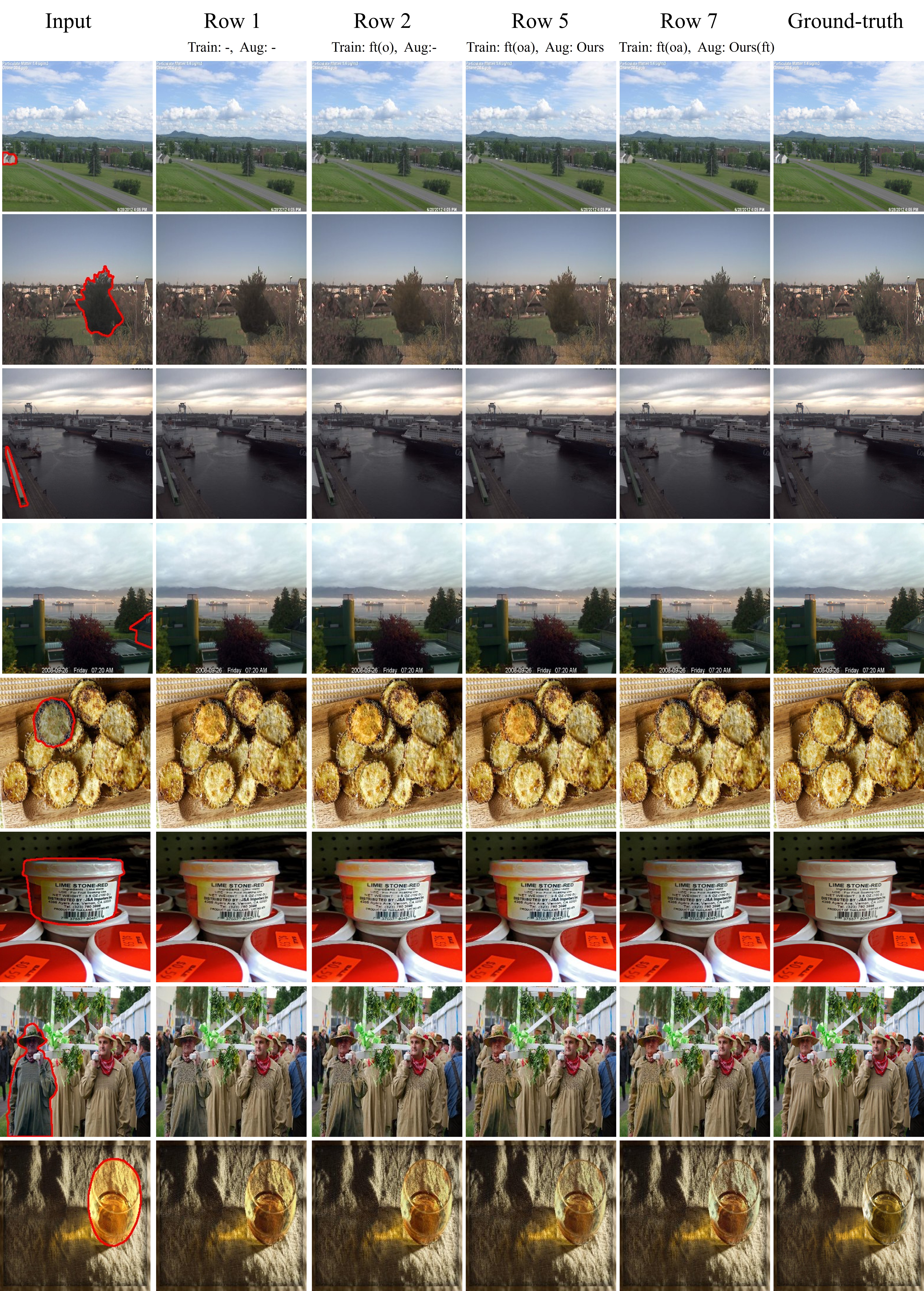} 
\caption{Examples of harmonized results on Hday2night (top 4 rows) and HFlickr (bottom 4 rows) datasets. The leftmost (\emph{resp.}, rightmost) column is the input composite image (\emph{resp.}, ground-truth real image), in which the foreground in the input image is outlined in red. The rest columns show the harmonized results of row 1, 2, 5, 7 in Table 1(a) of the main paper, in which row 7 is our full method. }
\label{fig:supp_examples_Hday2night_HFlickr}
\end{figure*}

Next, we show more harmonized results of different models (row 1, 2, 5, 7 in Table 1(a) of the main paper) in Figure \ref{fig:supp_examples_Hday2night_HFlickr} to supplement Figure 3 in the main paper.
Recall that row 1 is trained on the whole iHarmony4 training set, while row 2, 5, 7 are finetuned on Hday2night/HFlickr dataset. Row 2 does not use augmented training data, while row 5 and row 7 use augmented training data. The augmentation network used in row 5 is trained on iHarmony4, whereas the augmentation network used in row 7 is finetuned on each dataset. 

As shown in Figure \ref{fig:supp_examples_Hday2night_HFlickr},  the harmonized results from row 1 to row 7 are overall getting better. The results of row 7 are closest to the ground-truth real images, whereas the results of other rows may be insufficiently harmonized (\emph{e.g.}, ``house" in the first row) or adjusted towards an incorrect direction (\emph{e.g.}, ``cake" in the fifth row).  

\begin{figure*}
\centering
\includegraphics[width=0.9\textwidth]{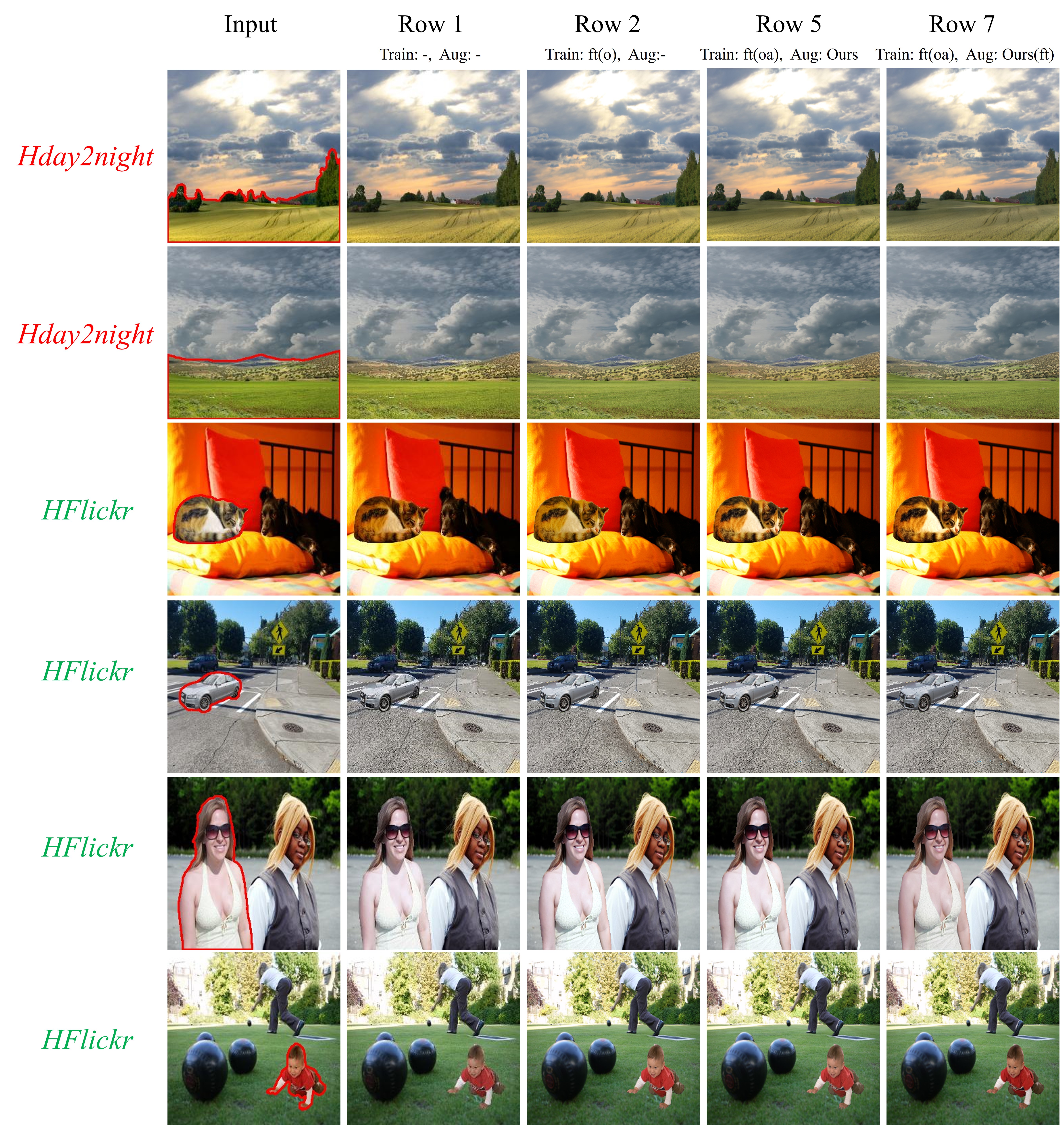} 
\caption{Examples of harmonized results for real composite images. The leftmost column is the input composite image, in which the foreground is outlined in red. The rest columns show the harmonized results of row 1, 2, 5, 7 in Table 1(a) of the main paper, in which row 7 is our full method. We also mark the domain (``Hday2night" or ``HFlickr") that each test composite image belongs to. }
\label{fig:supp_examples_real}
\end{figure*}

\begin{figure*}[t]
\centering
\includegraphics[width=0.9\textwidth]{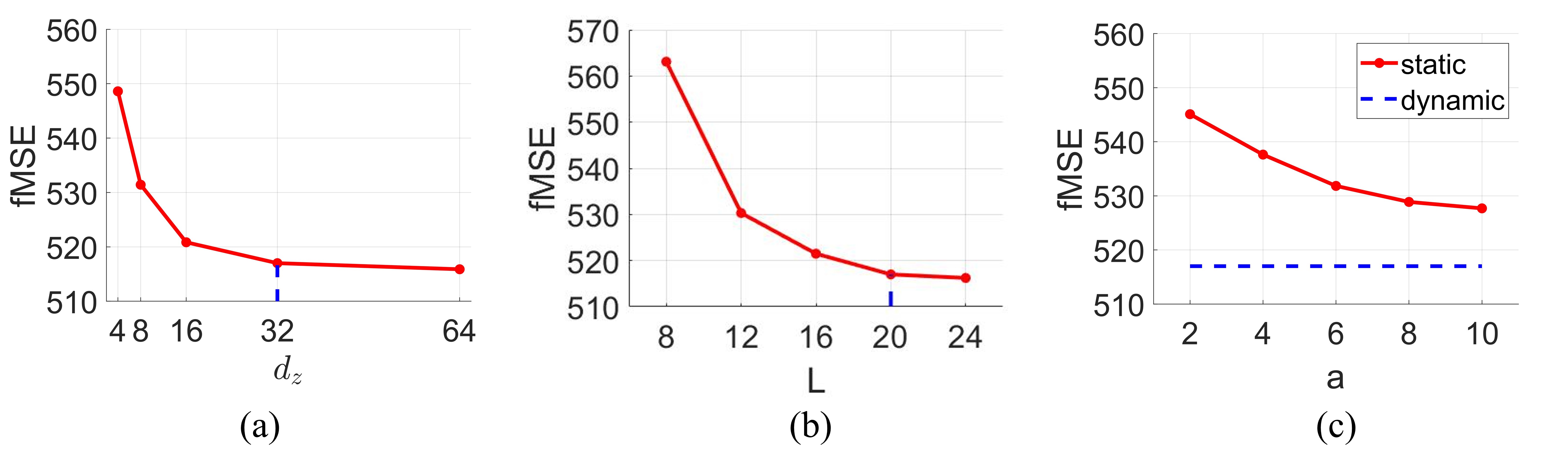} 
\caption{Figure (a)(b) show the performance variation when changing the dimension $d_z$ of latent code  and the number $L$ of basis LUTs, in which the vertical dashed line indicates the default value. Figure (c) compares static augmentation and dynamic augmentation when using different numbers of augmented images $N_a =a \times N$ for static augmentation.}
\label{fig:supp_hyper_parameter}
\end{figure*}

\section{Experiments on Large-scale Datasets} \label{sec:large_scale}
As mentioned in the main paper, we observe that the advantage of data augmentation is more obvious on small-scale datasets. In this section, we report the results on the other two relatively large datasets HCOCO and HAdobe5k from \cite{CongDoveNet2020} in Table \ref{tab:large_scale_datasets}. The results demonstrate that our method can achieve slight improvement on large-scale datasets. Based on the results on all four datasets (Table 1(a) in the main paper and Table \ref{tab:large_scale_datasets}), it can be seen that the results on large-scale datasets before using augmentation are better than those on small-scale datasets. The performance gain on large-scale datasets is relatively small, which indicates that the augmentation is more effective on small-scale datasets. 

\begin{table*}[t]
\centering
\begin{tabular}{|c|c|c|c|c|c|c|c|c|}
\hline
\multirow{2}{*}{} & \multirow{2}{*}{Train} & \multirow{2}{*}{Aug} & \multicolumn{3}{c|}{HCOCO} & \multicolumn{3}{c|}{HAdobe5k} \\  
\cline{4-9}
~ & ~ & ~ & MSE$\downarrow$ & fMSE$\downarrow$ & fSSIM $\uparrow$ &  MSE$\downarrow$ & fMSE$\downarrow$ & fSSIM $\uparrow$ \\ \hline
1 & - & - & 16.48 & 266.14 & 0.9224 & 22.59 & 166.19 &  0.9262 \\ 
\hline
2 & ft(o) & - & 15.67 & 257.34 & 0.9238 & 20.89 & 150.01  & 0.9310 \\ 
\hline
3 & ft(oa) & CT & 18.78 & 295.10  & 0.9154  & 25.99 & 187.58 & 0.9149 \\ 
\hline
4 & ft(oa) & LUT & 21.25  & 333.78 & 0.9146  & 31.48 & 220.07  &  0.9147 \\ 
\hline
5 & ft(oa) & Ours & 15.44 & 253.09 & 0.9237 & 19.39 & 143.78  & 0.9327  \\ 
\hline
6 & ft(a) & Ours(ft) & 15.98 & 261.23 & 0.9230  & 21.11 & 158.99  & 0.9275  \\ 
\hline
7 & ft(oa) & Ours(ft) & \textbf{15.25} & \textbf{251.11}  & \textbf{0.9239} & \textbf{18.59} & \textbf{132.32} & \textbf{0.9330}  \\ 
\hline
\end{tabular}
\caption{The experimental results using iS$^2$AM \cite{sofiiuk2021foreground} as image harmonization network on HCOCO and HAdobe5k. In ``Train" column, ``ft" means finetuning the harmonization model on Hday2night/HFlickr, and ``o" (\emph{resp.}, ``a") represents original (\emph{resp.}, augmented) training images. In ``Aug" column, ``LUT" (\emph{resp.}, ``CT") means using random LUT (\emph{resp.}, traditional color transfer methods) for data augmentation and ``Ours" means using our SycoNet for data augmentation. ``Ours(ft)" means finetuning SycoNet on HCOCO/HAdobe5k. The best results are highlighted in boldface.}
\label{tab:large_scale_datasets}
\end{table*}

\section{Evaluation on Real Composite Images} \label{sec:real_composite}

Following previous image harmonization works~\cite{tsai2017deep,CongDoveNet2020,ling2021region,sofiiuk2021foreground}, we also evaluate our method on real composite images. We collect 199 real composite images from \cite{tsai2017deep,cong2021high} (99 images from \cite{tsai2017deep} and 100 images from \cite{cong2021high}). 

As claimed in the main paper, the test composite image could be from any domain and we should use the harmonization model belonging to its closest domain. To find the matched domain for each real composite image, we train a domain classifier on the whole iHarmony4 training set, aiming to classify an image into four domains corresponding to four datasets. In detail, we concatenate the composite image and foreground mask as input, using ResNet50 \cite{he2015resnet} as the classification network.
We apply the trained domain classifier to the iHarmony4 test set and the classification accuracy is $95.48\%$. The high accuracy shows that the domain gap between four datasets is large enough to be separated by the trained domain classifier. Then, we apply the domain classifier to 199 real composite images, and acquire 10 (\emph{resp.}, 43) images belonging to Hday2night (\emph{resp.}, HFlickr) domain. 

We employ the models in row 1, 2, 5, 7 in Table 1(a) of the main paper. For the models which need to be finetuned on each domain (row 2, 5, 7), we employ the model from the matched domain for each real composite image. For example, we choose the model finetuned on ``Hday2night" for those real composite images classified as ``Hday2night" domain.  
Since there are no ground-truth harmonious images for these real composite images, we conduct user study for comparison. Following \cite{tsai2017deep,CongDoveNet2020}, given each composite image and its 4 harmonized outputs from 4 models, we can construct 10 image pairs by randomly selecting 2 from these 5 images. As a result, we can construct 530 image pairs based on 53 real composite images.
We ask 20 users to participate in this subjective evaluation. Each user could see an image pair each time and choose the one which looks more harmonious. In total, 20 users and 530 image pairs contribute to 10,600 pairwise results, based on which the Bradley-Terry (B-T) model \cite{bradley1952rank,lai2016comparative} is used to obtain the global ranking of all models. The B-T score of input composite image, row 1, row 2, row 5, and row 7 are -0.90, -0.04, 0.16, 0.30, and 0.47, respectively. Our finetuned augmentation network (row 7) achieves the highest B-T score. 
We also show the harmonized results in Figure \ref{fig:supp_examples_real}, \emph{i.e.}, two examples from ``Hday2night" domain and four examples from ``HFlickr" domain. From Figure \ref{fig:supp_examples_real}, we observe that the model with the assistance of our finetuned augmentation network (row 7) can generally produce more harmonious and realistic images, compared with other models.

\section{Comparison between Dynamic and Static Augmentation} \label{sec:cmp_dynamic_static}

As described in Section 3.2 in the main paper, we dynamically generate augmented composite images during the training procedure, which is dubbed as dynamic augmentation. There exists another straightforward augmentation strategy: generating augmented composite images beforehand and merging them into the original training set,  which is dubbed as static augmentation. For static augmentation, the number $N_a$ of augmented composite images needs to be decided. We set $N_a=a\times N$, in which $a$ varies in the range of $[2, 4, 6, 8, 10]$ and $N$ is the number of original training pairs. By taking Hday2night as an example, we plot the fMSE results of static augmentation with different $a$, and also plot the result of dynamic augmentation for comparison. Note that the experimental setting is the same as row 7 in Table 1(a) of the main paper. Static augmentation and dynamic augmentation use the same augmentation network.
As shown in Figure \ref{fig:supp_hyper_parameter}(c), the fMSE result of static augmentation first decreases and then begins to converge as $a$ increases. However, the best result achieved by static augmentation still lags behind dynamic augmentation. Moreover, dynamic augmentation generates augmented images on-the-fly without the need of storing the augmented images beforehand. Hence, dynamic augmentation is more elegant and effective than static augmentation.

\section{One Unified Model to Rule All Domains}\label{sec:unified_model}

In this section, we use learnable augmentation to enhance the unified harmonization model. Specifically, we use SycoNet pretrained on iHarmony4 to generate augmented images for the whole iHarmony4 training set. Then, we train the harmonization model iSSAM~\cite{sofiiuk2021foreground} on the augmented training set, and report the results (MSE/fMSE) on each test set (domain) in Table \ref{tab:unified_model}. 

\begin{table}[t] 
\centering
\label{tab:unified_model}
\resizebox{\columnwidth}{!}{
\begin{tabular}{|c|c|c|c|c|}
\hline
Dataset & HCOCO & HAdobe5k & HFlickr & Hday2night \\
\hline
w/o aug & 16.48/266.14 & 22.59/166.19 & 69.68/443.63 & 40.59/591.07 \\
with aug & 16.04/261.75 & 21.65/151.71 & 64.50/420.18 & 36.74/536.79 \\
\hline
\end{tabular}
}
\caption{The MSE/fMSE results on different test sets using a unified harmonization model with or without learnable augmentation.}
\end{table}

We still observe performance gain on each domain, which shows that our augmentation method can also improve the performance of a unified harmonization model on different domains. If the test sample is from a domain (\emph{e.g.}, HCOCO) with sufficient training data, the performance gain is small. If the test sample is from a domain (\emph{e.g.}, Hday2night) with limited training data, the performance gain is large. 

The results using one unified model are worse than the results using the model finetuned on each domain (Table 1(a) in the main paper and Table \ref{tab:large_scale_datasets}). Finetuning on each domain leads to multiple harmonization models for different domains, which is a little cumbersome, but the performance on each domain is higher.

\section{Hyper-parameter Analyses}\label{sec:hyper_parameter}

In our augmentation network, we have two hyper-parameters: the dimension $d_z$ of latent code and the number $L$ of basis LUTs. In this section, we investigate the impact of these two hyper-parameters by varying $d_z$ (\emph{resp.}, $L$) in the range of [4, 8, 16, 32, 64] (\emph{resp.}, [8, 12, 16, 20, 24]) while fixing the other one. The experimental setting is the same as row 7 in Table 1(a) of the main paper. By taking Hday2night as an example, we plot the fMSE results in Figure \ref{fig:supp_hyper_parameter}(a)(b), from which we see that fMSE first descends and then converges when $d_z$ or $L$ gets larger. Considering the trade-off between efficiency and effectiveness, we set $d_z=32$ and $L=20$ by default.

\begin{figure*}[t]
\centering
\includegraphics[width=0.99\textwidth]{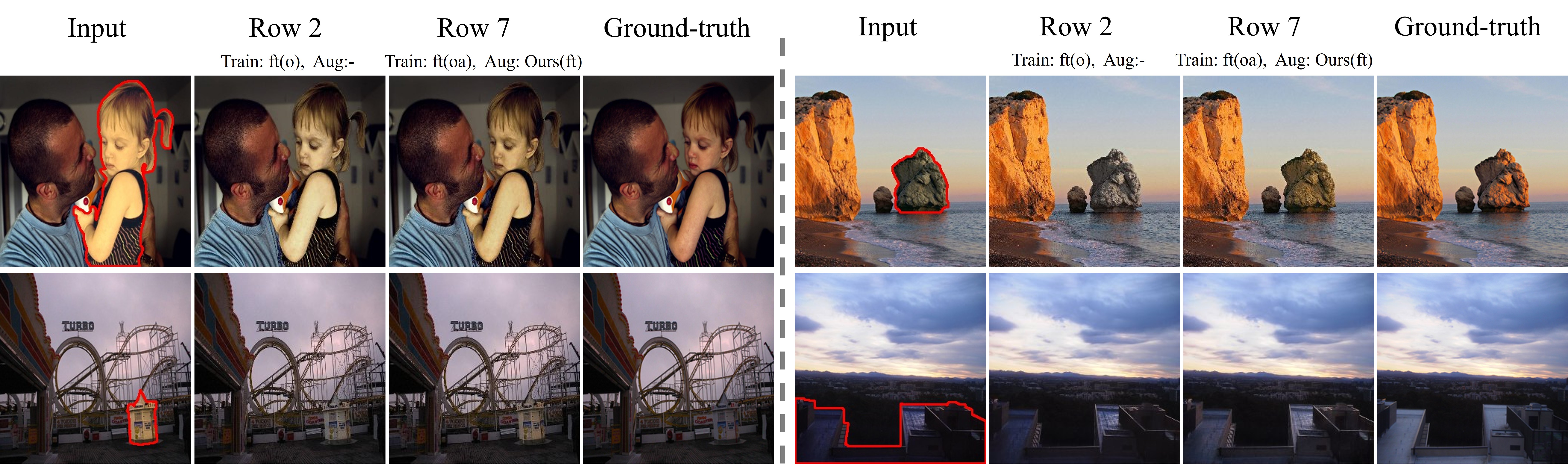} 
\caption{Four examples of failure cases. In each example, 
the first column is the input composite image with foreground outlined in red and the last column is the ground-truth real image. The second (\emph{resp.}, third) column is the harmonized result of row 2 (\emph{resp.}, 7) in Table 1(a) in the main paper, in which row 2 is the finetuned model without using augmentation and row 7 is our full method.}
\label{fig:failure_cases}
\end{figure*}

\section{Discussion on Limitation} \label{sec:limitation}
Although our learnable augmentation can usually enhance the performance, there still exist some unsatisfactory harmonization results. For examples, as shown in Figure \ref{fig:failure_cases},  the composite foreground has drastic illumination variation compared with real foreground or the background illumination condition is very complicated. In the first three examples, the foreground and background have dramatically different illumination statistics (\emph{e.g.}, brightness and color temperature). In the fourth example, the buildings in the background are very dark, but the foreground is relatively bright probably due to the sky light or other light sources. It may be arguable whether these images are hard cases from human perception, but they are indeed hard cases for image harmonization models. 
As shown in row 2, the basic image harmonization network without augmentation is struggling to harmonize these hard cases. The harmonized results are less harmonious and far from the ground-truth real image. The augmented images produced by our SycoNet fail to improve the ability of basic image harmonization networks to handle these hard cases (row 7). One possible reason is that these hard cases are relatively rare in the dataset, so our SycoNet does not fully cover these modes. One solution is pushing our SycoNet to produce more hard cases by assigning higher weights to the reconstruction losses of these hard cases.

\section{Computational Cost Analyses}\label{sec:computation_cost}
In the first training stage, we train the whole augmentation network SycoNet. The time of a feedforward pass (\emph{resp.}, FLOPs) is 0.0119s (\emph{resp.}, 11.95G).  
In the second training stage, we integrate the generation branch of SycoNet with a basic image harmonization network. By taking iSSAM as an example image harmonization network, the time of a feedforward pass (\emph{resp.}, FLOPs) of SycoNet generation branch is 0.0064s (\emph{resp.}, 7.32G), while the time of a feedforward pass (\emph{resp.}, FLOPs) of iSSAM is 0.0163s (\emph{resp.}, 15.04G). Thus, SycoNet generation branch is lightweighted compared with the basic image harmonization network.

More importantly, in the testing stage, we only use the basic harmonization network, so there is no extra computational cost.

\section*{Acknowledgement} The work was supported by the National Natural Science Foundation of China (Grant No. 62076162), the Shanghai Municipal Science and Technology Major/Key Project, China (Grant No. 2021SHZDZX0102, Grant No. 20511100300).

{\small
\bibliographystyle{ieee_fullname}
\bibliography{camera_supp.bbl}
}